\documentclass[10pt,twocolumn,letterpaper]{article}

\usepackage{cvpr}
\usepackage{times}
\usepackage{epsfig}
\usepackage{graphicx}
\usepackage{amsmath}
\usepackage{amssymb}
\usepackage{authblk}
\usepackage{adjustbox}
\makeatletter
\renewcommand\AB@affilsepx{\ \ \ \protect\Affilfont}
\makeatother
\usepackage{enumitem}

\newcommand{\JC}[1]{\textcolor{black}{{#1}}}

\cvprfinalcopy 

\def\cvprPaperID{1056} 

\ifcvprfinal\fi
\begin{document}

\title{Distort-and-Recover: Color Enhancement using Deep Reinforcement Learning}

\author[1]{Jongchan Park\thanks{This work was done while the author was at KAIST.}}
\author[2]{Joon-Young Lee}
\author[1,3]{Donggeun Yoo}
\author[3]{In So Kweon}
\affil[1]{\normalsize{Lunit Inc.}}
\affil[2]{Adobe Research}
\affil[3]{Korea Advanced Institute of Science and Technology (KAIST)}

\maketitle
\thispagestyle{empty}

\begin{abstract}
Learning-based color enhancement approaches typically learn to map from input images to retouched images. Most of existing methods require expensive pairs of input-retouched images or produce results in a non-interpretable way. In this paper, we present a deep reinforcement learning (DRL) based method for color enhancement to explicitly model the step-wise nature of human retouching process. We cast a color enhancement process as a Markov Decision Process where actions are defined as global color adjustment operations. Then we train our agent to learn the optimal global enhancement sequence of the actions. In addition, we present a `distort-and-recover' training scheme which only requires high-quality reference images for training instead of input and retouched image pairs. Given high-quality reference images, we distort the images' color distribution and form distorted-reference image pairs for training. Through extensive experiments, we show that our method produces decent enhancement results and our DRL approach is more suitable for the `distort-and-recover' training scheme than previous supervised approaches.
Supplementary material and code are available at \href{https://sites.google.com/view/distort-and-recover/}{https://sites.google.com/view/distort-and-recover/}
\end{abstract}

\begin{figure*}[t]
\setlength{\tabcolsep}{25pt}
\begin{center}
\includegraphics[width=0.8\textwidth]{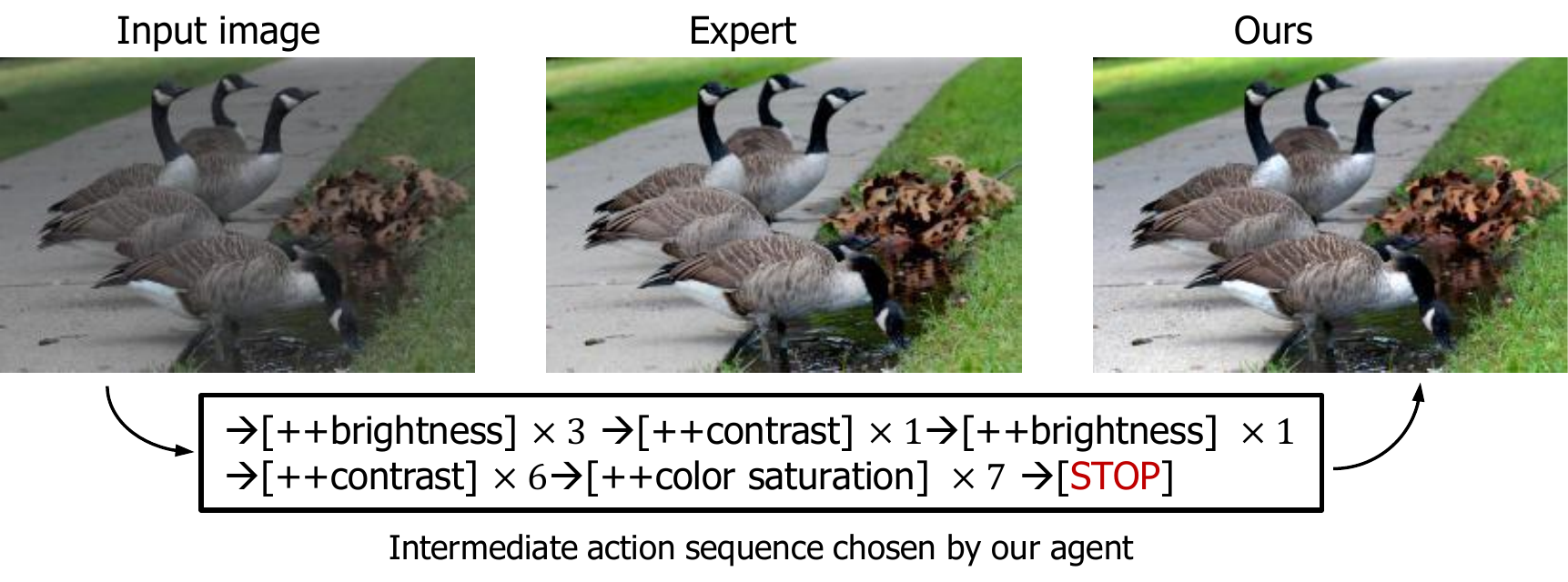}
\end{center}
\vspace{-4mm}
\caption{One example of our method. Using deep reinforcement learning, we train our agent with human expert's images in MIT-Adobe FiveK dataset~\cite{fivek}. The agent iteratively selects an editing operation to apply and automatically produces a retouched image with an interpretable action sequence. }
\label{fig:example}
\vspace{-2mm}
\end{figure*}
\section{Introduction}

With the widespread use of digital imaging devices, photo retouching is becoming more popular. Professional photo retouching software such as Adobe Photoshop or Lightroom provides various retouching operations, but it requires expertise in photo editing and also lots of effort for achieving satisfying results. Casual software, such as Instagram on mobile platforms, provides several predefined stylization filters which make photo retouching easier for users, but the filters only perform predefined operations regardless of the context or mood of a photo and result in poor results in many cases.

Automatic color enhancement is a non-trivial task because of the highly non-linear and subjective nature of photo retouching.
Consider a human professional, the retouching procedure is a sequence of iterative decision making given the image's context and color distribution.
A human professional iteratively applies retouching operations until the color distribution fits the individual's taste.
Therefore, the translation of pixel values to the optimal states is a complex combination of the pixel's value and the global/local color/contextual information.
Also, it is a perceptual process such that there can be varying optimal states for different individuals~\cite{fivek,lee2016automatic}.
As a result, the translation from an input color distribution to the optimal becomes highly non-linear and multi-modal.

There are several lines of research for automatic color enhancement.
Exemplar-based methods~\cite{Hwang2012,liu2014autostyle,lee2016automatic} tackle the non-linearity and/or multi-modality by transferring exemplar images' color distribution to the target image. During the process, exemplar images are given or retrieved from a database and the target image's color distribution is translated into the exemplar images' color distribution. 
In the learning-based approaches~\cite{fivek,yan2014learning,Yan2016}, a mapping function from the source color distribution to the target color distribution is learned from training data. Recent work~\cite{Yan2016} handles the high non-linearity of this problem with a deep neural network. The multi-modality problem is not yet explicitly solved by any learning-based method. 
A general problem of learning-based methods is expensive paired datasets, such as MIT-Adobe FiveK dataset~\cite{fivek} of input-retouched image pairs or the dataset~\cite{yan2014learning} of input-action pairs.

In this paper, we propose a novel approach for automatic color enhancement with two distinctive aspects. First, we present a deep reinforcement learning (DRL) approach for color enhancement (see Figure~\ref{fig:example}). We cast the problem into a Markov Decision Process (MDP) where each step action is defined as a global color adjustment operation, such as brightness, contrast, or white-balance changes. Therefore it explicitly models iterative, step-by-step human retouching process. We solve such MDP problem with a well-studied deep reinforcement learning framework, Deep Q-Network (DQN)~\cite{Mnih2015}.

Second, we propose an economic \textit{`distort-and-recover'} training scheme to avoid using expensive paired datasets. It is a simple and straightforward approach that only requires a set of high-quality reference images to learn color enhancement. We randomly distort the collected reference images and form distorted-reference image pairs for training. Such distorted-reference dataset is more economic than an input-retouched image set because high-quality reference images can easily be collected in personal photo albums or stock image websites. 
In the experiment section, we show that the \textit{distort-and-recover} scheme is suitable for the DRL approach due to the state-exploring nature.



There are several potential applications with our method. Thanks to the economic training data collection, it is possible to train a personalized retouching agent using images retouched or selected by a user. Our experiment shows that the agents learn different styles with training datasets with different styles. In addition, if the retouching operations in our agent are defined as the operations in retouching software, then our agent can be seamlessly integrated into the retouching software. 


\paragraph{Contributions.} Our main contribution is three-fold.
\begin{enumerate}[topsep=0pt,itemsep=0pt]
\item We present a color enhancement agent that learns a step-by-step retouching process without any explicit supervision for intermediate steps.
\item We propose a \textit{distort-and-recover} training scheme that enables us to train our agent without expensive input-retouched image pairs.
\item We show that our agent, trained with the distort-and-recover scheme, can enhance images from unknown color distributions. Through a user study, we show our agent outperforms competitive baseline methods including a recent supervised learning algorithm and an auto-retouching algorithm in a commercial software.
\end{enumerate}

\section{Related Work}
\label{sec:related}
One traditional approach for color enhancement is transferring the color of an example image to a given input image. It is originated from \cite{Reinhard:2001:CTI:616072.618848} in which the global color distribution of an input image is warped to mimic an example style. There are many subsequent works to improve this technique~\cite{faridul2014survey}. While this approach can provide expressive enhancement and diverse stylizations, the results highly depend on example images while providing proper exemplars is challenging. Recent works~\cite{liu2014autostyle,lee2016automatic,Hwang2012} (semi-)automate exemplar selection by image retrieval methods. Liu~\etal\cite{liu2014autostyle} used a keyword-based image search to choose example images. Lee~\etal\cite{lee2016automatic} learn a content-specific style ranking using a large photo collection and select the best exemplar images for color enhancement. For pixel-wise local enhancement, Hwang~\etal\cite{Hwang2012} find candidate images from a database then search local color enhancement operators. 

Learning-based color enhancement is another dominant stream~\cite{fivek,yan2014learning,Yan2016}. Bychkovsky~\etal\cite{fivek} present a number of input and retouched image pairs called MIT-Adobe FiveK, which is created by professional experts. They used this data to train a model for color and tone adjustment. Yan~\etal\cite{Yan2016} propose a deep learning method to learn specific enhancement styles. Given the features of color and semantic context, a deep neural network as a non-linear mapping function is trained to produce the pixel color of specific styles.

In terms of step-by-step retouching process modeling, the work in \cite{yan2014learning} is the closest to our work. In~\cite{yan2014learning}, Yan~\etal used a learning-to-rank approach for color enhancement. They collected the intermediate editing actions taken in the retouching process and used this data to learn a ranking model that evaluates various color enhancements of an image. In the inference stage, this method takes various actions and evaluates the results with the ranking model to select the best action. This process is repeated to obtain a final result. There are many differences between Yan~\etal\cite{yan2014learning} and our method. Our method is much efficient as our agent directly selects an optimal action for each step instead of evaluating various action trials. More importantly, our method does not require step-wise action annotations, which are prohibitively expensive. Also, the action set in \cite{yan2014learning} is limited by annotated datasets while our action set does not depend on the dataset. In practice, most existing learning-based methods require at least input-retouched pairs of images for training and learn to map from known source distributions to known target distributions. The difficulty in paired data collection makes personalized enhancement difficult. Also existing datasets have strong bias to specific input distributions (\eg input images in MIT-Adobe FiveK), leading to poor generalization performance. On the other hand, we show that our method can be trained solely with reference images or retouched images and learn to map to target distributions from unknown input distributions. 

Recently, generative adversarial networks show impressive performance on many challenging tasks. Using such network, Pix2Pix~\cite{Pix2Pix} translates an image to other image and has shown strong potential to the color enhancement application. It turns day images to night images or gray images to color images, and we believe this network can be applied to the color enhancement task where input images are transformed to retouched images. In the experiment, we show that Pix2Pix achieves color enhancement performance close to the state-of-the-art~\cite{Yan2016}. Setting Pix2Pix as a baseline, our method outperforms Pix2Pix under the `\textit{distort-and-recover}' training scheme according to our user study results in Section~\ref{sec:exp_distort_and_recover}.

\section{Problem Formulation}
\label{sec:formulation}
A human expert enhances images by applying a set of color adjustment operations. To imitate the process, we formulate color enhancement as a problem for finding an optimal sequence of adjustment actions as follows.

We enhance an input image $\mathcal{I}$ by iteratively applying an adjustment action $\mathcal{A}$. We represent an image $\mathcal{I}(t)$ at a step $t$ using a contextual feature $\mathcal{F_\text{context}}(I\text(t))$ and a global color feature $\mathcal{F_\text{color}}(I\text(t))$. 
Our agent determines a color adjustment action $\mathcal{A}\text(t)$ at each step $t$ under the policy $\Omega_\Theta$. Therefore, our goal is to find an optimal sequence of color adjustment actions $\mathcal{T}\{\mathcal{A_\text{optimal}}\text(t)\subset\boldsymbol{\mathcal{A}}\}$ that gives the best improvement of image color.

To solve this problem, we need a metric to measure the color aesthetic of an image $\mathcal{I}(t)$. The color aesthetic depends on not only the semantic context of the image but also individual preference, therefore it is difficult to define an absolute aesthetic score function. There has been a long line of works to assess the aesthetic scores of images as summarized in~\cite{DengLT16}, but there is still no winning method that gives satisfying results. Recently, deep learning based works~\cite{lu2014rapid,lu2015deep,kong2016photo} lead large improvements on the problem, so we initially considered representative networks as our color aesthetic metric but the results were not stable enough for our test images. Instead, to measure the color aesthetic of an image, we simply consider human retouched images $\mathcal{I_\text{target}}$ as ground truth and define our aesthetic metric as a negative $\ell_2$ distance between an image $\mathcal{I}(t)$ and corresponding retouched image $\mathcal{I_\text{target}}$, which is also used in~\cite{Yan2016,Hwang2012}. 
Now, our goal is to find an optimal sequence of color adjustment actions $\mathcal{T}\{\mathcal{A_\text{optimal}}\text(t)\subset\boldsymbol{\mathcal{A}}\}$ that minimizes $\left\|\mathcal{I}(t_{final})-\mathcal{I_\text{target}}\right\|^2$.

This formulation can be regarded as a Markov Decision Process, where the state $\mathcal{S}$ is a combination of the contextual feature and the color feature $(\mathcal{F_\text{context}},\mathcal{F_\text{color}})$, the action space is the set of color adjustment operations $\boldsymbol{\mathcal{A}}$, and the immediate reward is the change in $\ell_2$ distance as:
\begin{equation}
	\mathcal{R}\text(t)=\left\|\mathcal{I_\text{target}}-\mathcal{I}\text(t-1)\right\|^2-\left\|\mathcal{I_\text{target}}-\mathcal{I}(t)\right\|^2.
	\label{eq:reward}
\end{equation}
An agent, parameterized by $\Theta$, determines the policy $\Omega_\Theta$. Unlike the typical Markov Decision Process setups, there is no transitional probabilities since the operation for each color adjustment action is deterministic.

To solve this decision-making problem, we train the agent to approximate the action value $\mathcal{Q}(\mathcal{S}(t),\mathcal{A})$ and choose an action $\mathcal{A}$ that maximizes the value. The action value of an action $\mathcal{A}$ at time $t$ is an expected sum of future rewards as:
\begin{equation}
	\mathcal{Q}(\mathcal{S}(t),\mathcal{A})=E\left[r(t)+\gamma\cdot r(t+1)+\gamma^2\cdot r(t+2)+\cdots\right], \nonumber
\end{equation}
where $\gamma$ is a discount factor. The state space, composed of the context feature and the color feature, is continuous so the complexity of the approximation is high. We therefore employ a deep neural network as an agent to handle such a high complexity. This agent estimates the action value $ \mathcal{Q}(\mathcal{A}, t) $ for $ \mathcal{S}\text(t) $. 

\begin{figure*}
	\centering
	\includegraphics[width=0.9\textwidth]{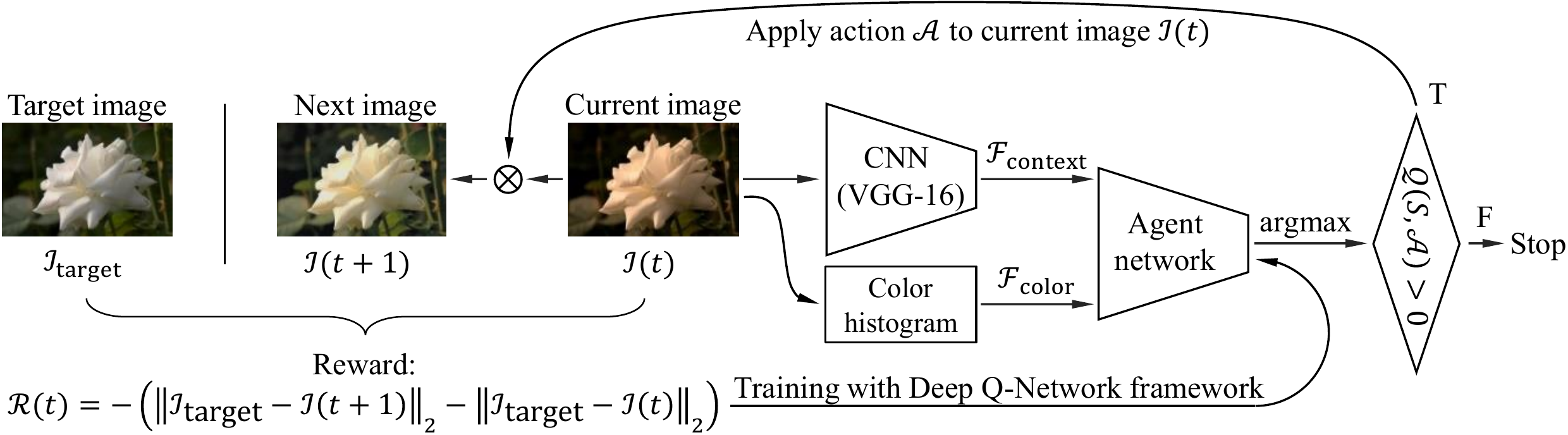}
	\caption{\textbf{Overview of our step-wise automatic color enhancement method.} Given an image $\mathcal{I}(t)$ at a sequential adjustment step $t$, we extract a contextual feature $\mathcal{F_\text{context}}$ with a pre-trained CNN and a color feature $\mathcal{F_\text{color}}$. These features are forwarded to the agent network, then the agent determines an optimal action $\mathcal{A}$ to make the image look better. We apply the action to the current image, and repeat this process until the agent produces a ``stop'' signal. Under the Deep Q-Network framework, this agent is trained to maximize the reward defined as a pixel-level distance between an input and a target retouched image.}
	\label{fig:architecture}
\end{figure*}


\section{Automatic Color Enhancement}
\label{sec:architecture}

Figure~\ref{fig:architecture} illustrates the overall pipeline of our automatic color enhancement algorithm. To enhance the color of an input image, our method proceeds as follows. An input image $\mathcal{I}$ is forwarded to the feature extractor to extract features $\mathcal{S}$=($\mathcal{F_\text{context}}, \mathcal{F_\text{color}}$). The agent then takes the features and estimates the action value $\mathcal{Q}(\mathcal{S},\mathcal{A})$ for each predefined adjustment action $\mathcal{A}$. The agent then chooses an action that has the highest action value and applies the action to the input image if the value is positive.
The agent repeats this process and stops when all estimated action values are negative.


\subsection{Actions}
\label{sec:actions}
As commercial photo retouching software provides various built-in actions, we also define several editing actions as shown in Table~\ref{table:actions}. We take a part of actions in~\cite{yan2014learning} and add additional actions for white-balancing. In our action items, Actions of 1 and 2 are for adjusting image contrast, and Actions of 3 and 4 are for controlling color saturation. Image brightness is controlled by Actions of 5 and 6. For white-balancing, we define Actions from 7 to 12, which increase or decrease two color components. The actions are quantized since DQN~\cite{Mnih2015} accepts discrete actions.


\begin{table}[t]
	\begin{center}
		\setlength{\tabcolsep}{3.5pt}
		\small
		\begin{tabular}{|c|l||c|l|}
			\hline
			\#&Action description&\#&Action description\\ \hline\hline
			1&$\downarrow$ contrast ($\times$0.95)&
			2&$\uparrow$ contrast ($\times$1.05)\\\hline
			3&$\downarrow$ color saturation ($\times$0.95)&
			4&$\uparrow$ color saturation ($\times$1.05)\\\hline
			5&$\downarrow$ brightness ($\times$0.95)&
			6&$\uparrow$ brightness ($\times$1.05)\\\hline
			7&$\downarrow$ red and green ($\times$0.95)&
			8&$\uparrow$ red and green ($\times$1.05)\\
			9&$\downarrow$ green and blue ($\times$0.95)&
			10&$\uparrow$ green and blue ($\times$1.05)\\
			11&$\downarrow$ red and blue ($\times$0.95)&
			12&$\uparrow$ red and blue ($\times$1.05)\\\hline
		\end{tabular}
	\end{center}
	\caption{\textbf{Actions for automatic color enhancement.} We define 12 actions to adjust contrast, saturation, brightness, and whit-balance. Each action increases or decreases the value by 5\%.}
	\label{table:actions}
	\vspace{-2mm}
\end{table}

\subsection{Features}
\label{sec:features}
The agent determines which action to take based on information embedded in features. It is important to define features with information that a human considers for photo retouching, such as color distribution and semantic context. 
A person's preferred color distribution of an image highly depends on the semantic context of an image, therefore contextual information is an important cue to determine the color of an image. For this reason, Hwang~\etal\cite{Hwang2012} adopt GIST~\cite{torralba2003contextual} as a contextual feature, Bychkovsky~\etal\cite{fivek} utilize face detection results to define a contextual feature, and Yan~\etal\cite{Yan2016} utilize a semantic label map estimated from scene parsing and object detection algorithms. However, GIST is not sufficient to encode high-level semantic information and relying on separate algorithms for object detection or segmentation is not efficient.

Instead, we utilize the intermediate activations of a deep convolutional neural network pre-trained on the ILSVRC classification dataset~\cite{russakovsky2015imagenet}. Since the network is trained to recognize 1,000 object classes with 1.3M images, intermediate activations of such network embed semantic information of an image. As studied by \cite{sharif2014cnn,yoo2015multi}, the use of the activations as a generic image feature significantly improves other visual recognition tasks. We choose the 4,096-dimensional activations from the sixth layer of VGG-16 model~\cite{Simonyan14c} as our contextual feature $\mathcal{F}_\text{context}$.

For color features $\mathcal{F}_\text{color}$, we adopt a CIELab color histogram. We linearly quantize each axis of the CIELab space to 20 intervals and count the number of pixels fall into each interval to obtain a 20$\times$20$\times$20-dimensional histogram. In addition to this feature, we have tried various types of color features, but CIELab histogram shows the best performance (see Section~\ref{sec:exp_feature}).


\subsection{Agent}
\label{sec:agent}

The agent takes the features extracted from the current image and estimates the action value for each of actions. The action value $\mathcal{Q}(\mathcal{S}(t),\mathcal{A})$ of an action $\mathcal{A}$ is the expected sum of future rewards with the action. 
Our agent consists of a 4-layer multi-layer perceptron and all layers use a rectified linear unit as activation functions.

During training, the policy $\Omega_\Theta$ is determined with an $\epsilon$-greedy algorithm. The $\epsilon$-greedy algorithm randomly samples actions with a probability of $\epsilon$ and chooses the actions with the highest expected return in a greedy manner with a probability of $1-\epsilon$. The agent is trained with the intermediate reward $\mathcal{R}(t)$ of Equation~\ref{eq:reward} which is defined as the change of the negative $\ell_2$ distance. In the inference stage, we determine the policy with a pure greedy algorithm where $\epsilon$ is zero, \textit{i.e.} the highest expected return is always chosen.
The process is repeated until all expected returns are negative.

\section{`Distort-and-Recover' Training Scheme}
\label{sec:data}
Despite the successful color enhancement results reported in~\cite{fivek,yan2014learning,Yan2016}, the learning-based methods are heavily dependent on datasets by nature. All the previous learning-based approaches require at least the input-retouched image pairs for training. Paired datasets are expensive in terms of data collection, thus it makes difficult to develop a system to account for personalized image enhancement. More importantly, such datasets often cover specific input distributions only and it leads to poor generalization due to large distribution changes of inputs (\eg when an iPhone image is tested with a color retouching model trained with raw input images on MIT-Adobe FiveK~\cite{fivek}).

To resolve the problem, we propose a `\textit{distort-and-recover}' training scheme, which only utilizes retouched or curated reference images.
We distort high-quality reference images by randomly applying color adjustment operations and synthesize a \textit{pseudo} input-retouched pairs of images.
In order to deliver clearer supervisory signals with efficient search space, the $\mathcal{L}^2$ distance from the distorted image to the reference image is kept from 10 to 20 in the CIELab colorspace.
\JC{Also, to prevent the bias of color distortion, we use different global operations from the DRL agent's action set: brightness/contrast/color saturation adjustments in highlight/shadow pixels, and C/M/Y/R/G/B adjustments in highlight C/M/Y/R/G/B pixels respectively. In selecting highlight/shadow pixels, we use a soft pixel selection method with a variant of sigmoid function which applies high weights on pixels with high/low values. Other than the highly non-linear operations, we also use basic brightness/contrast/color saturation adjustments. Distortion operations are designed with simplicity in mind. Any type of global operations can be used for distortion.}
Details about the random distortion operations are shown in the supplementary materials.

One may argue that there is expensive human labor included in the retouched reference images, but in the aspect of data collection, retouched reference images are already openly available in many areas. 
For example, high-quality reference images can be collected from stock image websites, Flickr, or Aesthetic Visual Analysis dataset~\cite{ava}.
Input-retouched image pairs, however, are not usually openly available thus creating a new paired dataset requires intensive labor of professionals.

We will validate the effectiveness of the `\textit{distort-and-recover}' training scheme in Section~\ref{sec:exp_distort_and_recover}. Surprisingly, our DRL agent works well with the synthesized pairs while a supervised learning method performs poorly.

\section{Experiments}
\label{sec:experiment}

In this section, we validate our method through extensive experiments: 1) evaluation with different sets of features as input to the agent network in Section~\ref{sec:exp_feature}, 2) evaluation on input-retouched paired datasets in Section~\ref{sec:exp_raw_retouched}), 3) evaluation of the ``\textit{distort-and-recover}" training scheme with a user study in Section~\ref{sec:exp_distort_and_recover}.

\subsection{Experimental Setup}
We have built our network with Tensorflow\footnote{https://www.tensorflow.org/}. 
Among various follow-up studies of DQN, we adopt the Double Q-Learning method~\cite{van2016deep} for deep reinforcement learning.
Our agent network consists of 4 fully-connected layers, the output size of the layers are {4096, 4096, 512, 12}, and the input size may vary depending on the type of input features in use.
It takes at least 12 hours to train the agent network on a single NVIDIA GTX 1080 with MIT-Adobe FiveK dataset. The duration varies depending on the size of training set.
The mini-batch size is 4, the base learning rate is $10^{-5}$, the minimum learning rate is $10^{-8}$, and learning rate decays by a factor of 0.96 every 5,000 iterations. Among many default optimizers provided in Tensorflow, for our task, Adam optimizer shows the most stable training with the best result.

\paragraph{Pix2Pix Baseline.}
The current state-of-the-art method in automatic color enhancement is \cite{Yan2016}, but it is hard to directly compare results\footnote{The main code of \cite{Yan2016} is available, but we fail to get reasonable results due to the external dependencies on core modules such as the semantic label map. The full set of final results is not openly available.}. Instead, we use Pix2Pix~\cite{Pix2Pix} as a strong baseline for a pixel-level prediction method. Pix2Pix is a conditional generative adversarial network that translates images into a different domain, and we believe it is a general and advanced technique for image generation. Thus, Pix2Pix is applicable to the color enhancement problem. 

For fair comparisons, we tune the hyperparameters of Pix2Pix throughout extensive trial and errors to get the best enhancement results.
The input of the Pix2Pix implementation~\footnote{https://github.com/affinelayer/Pix2Pix-tensorflow} is $256\times256$ squared images. Direct warping to squared images may distort the color distribution of the images, so we keep the aspect ratio and add zero-pixels to form squared input-output pairs and then resize them into $256\times256$ resolution.
The input channels are changed from the RGB to CIELab colorspace in which pixels are re-scaled to [-1, +1], and the RGB $\mathcal{L}^1$ loss is changed to the CIELab $\mathcal{L}^2$ loss.


\paragraph{Dataset and metrics.} 
MIT-Adobe FiveK dataset~\cite{fivek} is mainly used for our experiments. It consists of 5,000 raw images, each of which has paired with retouched images from five different experts -- A/B/C/D/E, so there are five sets of 5,000 input-retouched paired images. If not otherwise specified, as our test set, we use raw input images on \textsc{Random 250} which is a subset of MIT-Adobe FiveK dataset randomly selected in ~\cite{Hwang2012}, and the rest of 4,750 images are used for training.
For quantitative comparison, we use the mean $\mathcal{L}^2$ error metric to evaluate the performance, following the previous approaches ~\cite{Hwang2012,Yan2016}.

\begin{table}[t]
\begin{center}
\setlength{\tabcolsep}{1.7pt}\small
\begin{tabular}{|l|c|}
\hline
Features&mean $\mathcal{L}^2$ error\\\hline\hline
Context + RGB-L histogram& 12.62\\
Context + TinyImage& 15.07\\
Context + Six features in \cite{fivek}& 12.53\\
Context + CIELab histogram&\textbf{10.99}\\
Lab histogram&12.30\\\hline
\end{tabular}
\end{center}
\vspace{-2mm}
\caption{\textbf{Comparison of mean $\mathcal{L}^2$ error} with different feature combinations on \textsc{Random 250}. We use six features as in \cite{fivek}.
}
\label{table:f_comparison}
\vspace{-2mm}
\end{table}

\begin{table}
	\begin{center}
		\setlength{\tabcolsep}{1.7pt}\small
		\begin{tabular}{|l|c|c|}
			\hline
			Method&mean $\mathcal{L}^2$ error&SSIM\\\hline\hline
			Input image&17.07&-\\
			Exemplar-based (Hwang~\etal\cite{Hwang2012})&15.01&-\\
			DeepNet-based (Yan~\etal\cite{Yan2016}) &\textbf{9.85}&-\\
			Pix2Pix~\cite{Pix2Pix}&10.49&0.857\\
			Ours&10.99&\textbf{0.905}\\\hline
		\end{tabular}
	\end{center}
	\vspace{-2mm}
	\caption{\textbf{Comparison of mean $\mathcal{L}^2$ error} in different approaches on \textsc{Random 250}. All approaches are trained with input-retouched image pairs by Expert C on MIT-Adobe FiveK.}
	\label{table:comparison1}
	\vspace{-5mm}
\end{table}

\subsection{Feature Selection}
\label{sec:exp_feature}
\JC{In this section, we compare the results from various combinations of context and color features as the input to the agent network.}
As described in Section~\ref{sec:features}, we use the convolutional feature at the sixth layer of VGG16~\cite{Simonyan14c} model pretrained on ImageNet~\cite{russakovsky2015imagenet} as our contextual feature. Together with the contextual feature, we evaluate various color features, since deep features contain high-level semantic information while the low-level information is important for color enhancement.

We train the agent network with different sets of features. We use the input-retouched image pairs by Expert C on MIT-Adobe FiveK and measure the mean $\mathcal{L}^2$ error for quantitative comparison. All other setups are identical.

Table~\ref{table:f_comparison} demonstrates that the combination of the CIELab color histogram and the contextual feature yields the best performance in terms of mean $\mathcal{L}^2$ error on \textsc{Random 250}. It is also verified that the contextual feature plays an important role for color enhancement as the error increases a lot without the contextual feature. All experiments in the following sections, we use the best feature combination, the VGG feature and the CIELab color histogram.

\begin{figure*}
\begin{center}
\begin{adjustbox}{max width=\textwidth}
\setlength{\tabcolsep}{1pt}\footnotesize
\begin{tabular}{cccccc}
\includegraphics[width=0.17\linewidth]{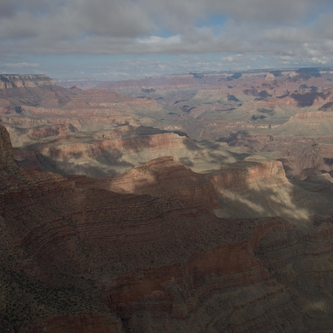}&
\includegraphics[width=0.17\linewidth]{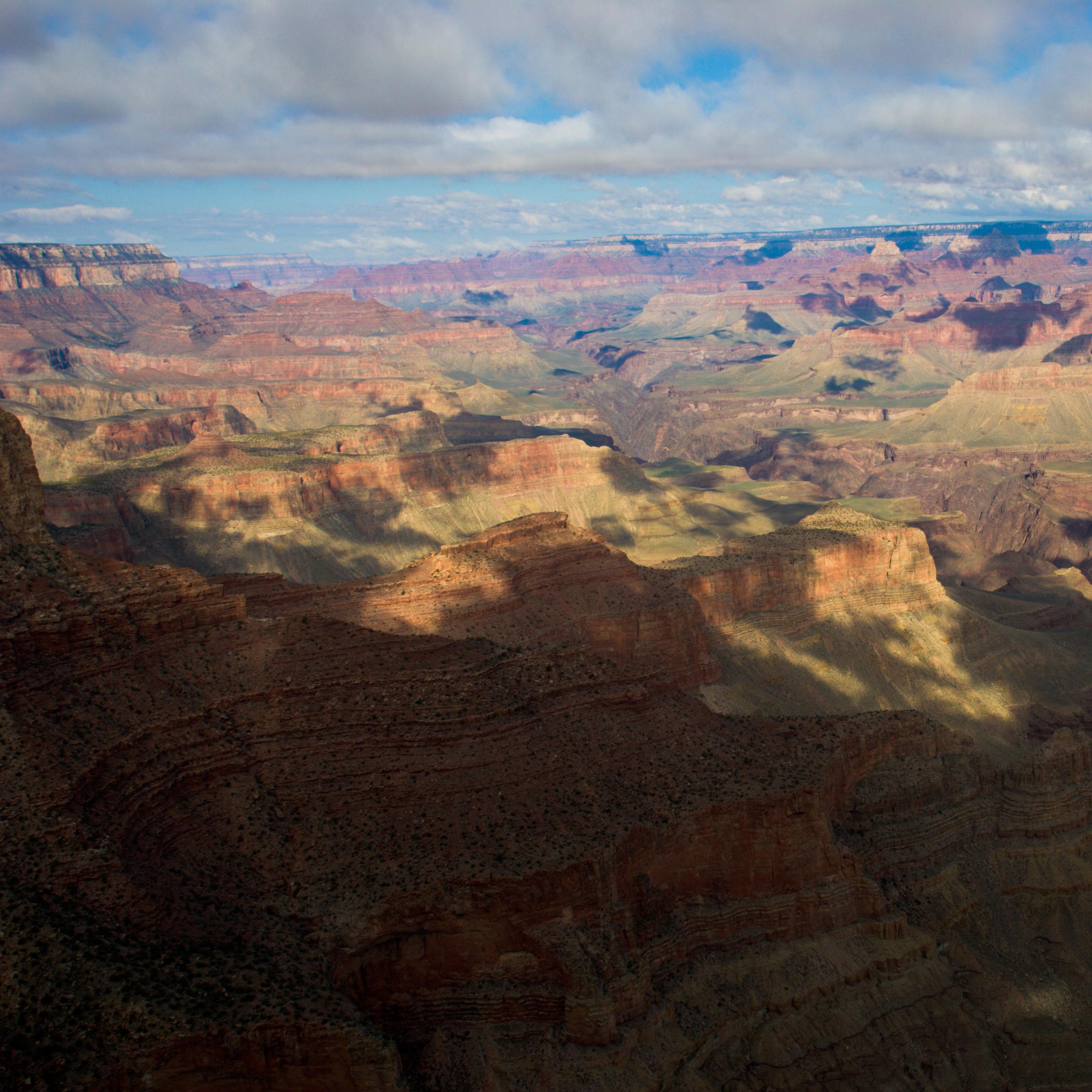}&
\includegraphics[width=0.17\linewidth]{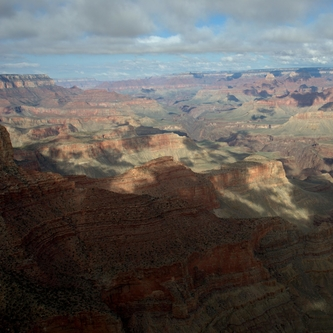}&
\includegraphics[width=0.17\linewidth]{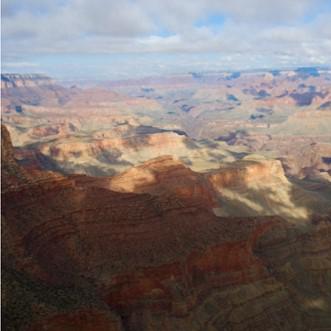}&
\includegraphics[width=0.17\linewidth]{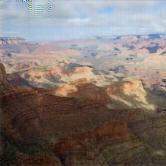}&
\includegraphics[width=0.17\linewidth]{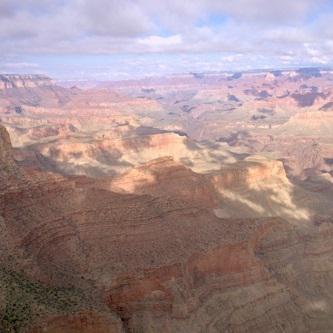}\\
\includegraphics[width=0.17\linewidth]{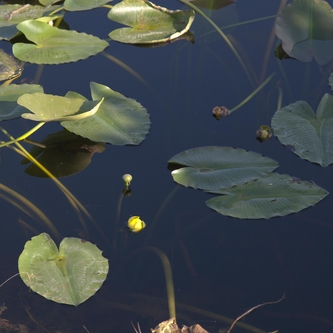}&
\includegraphics[width=0.17\linewidth]{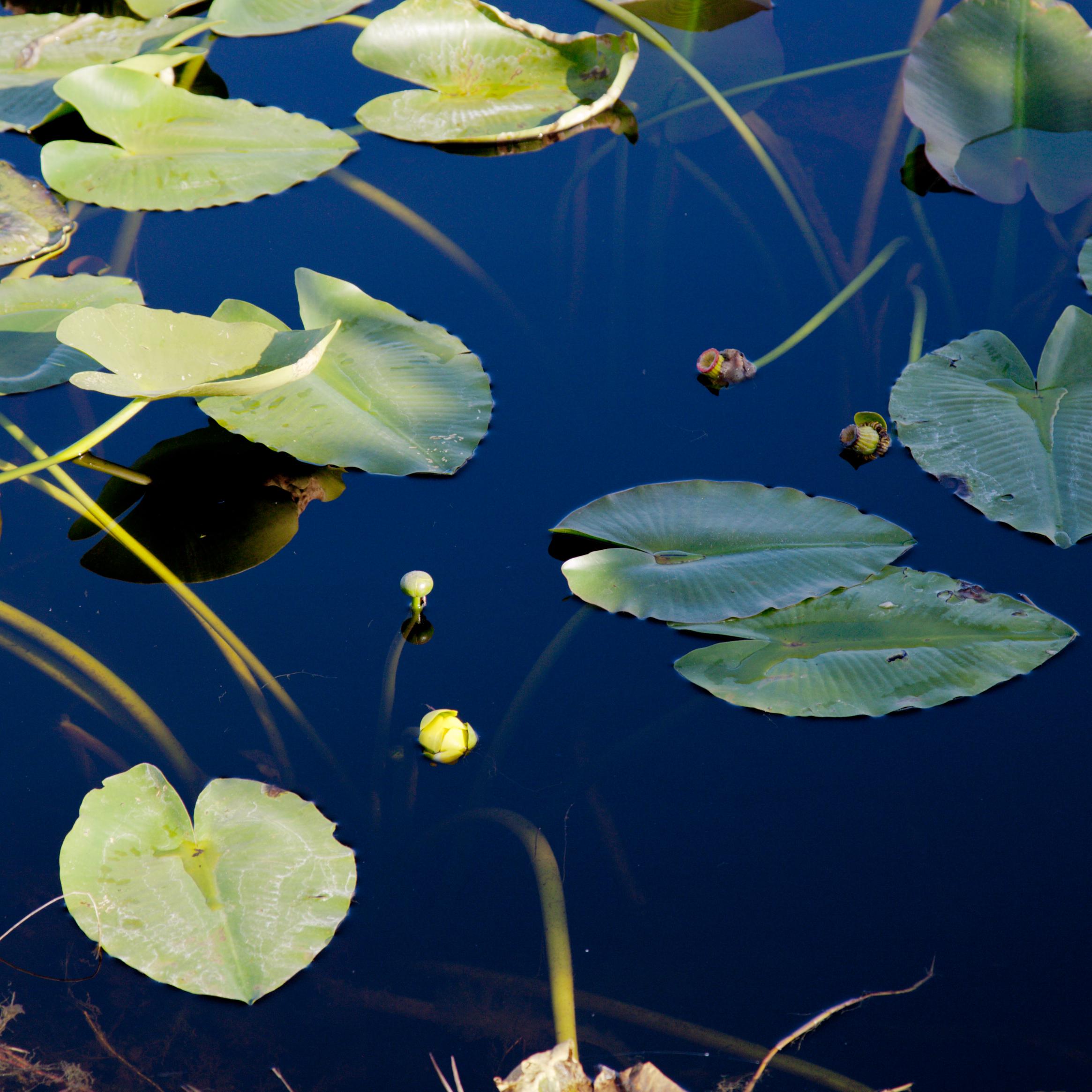}&
\includegraphics[width=0.17\linewidth]{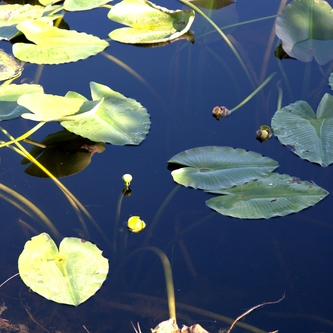}&
\includegraphics[width=0.17\linewidth]{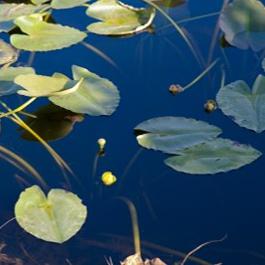}&
\includegraphics[width=0.17\linewidth]{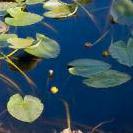}&
\includegraphics[width=0.17\linewidth]{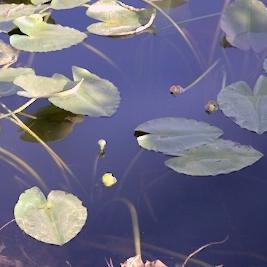}\\
\includegraphics[width=0.17\linewidth]{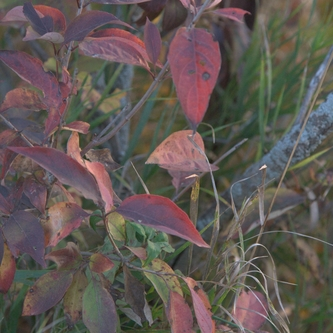}&
\includegraphics[width=0.17\linewidth]{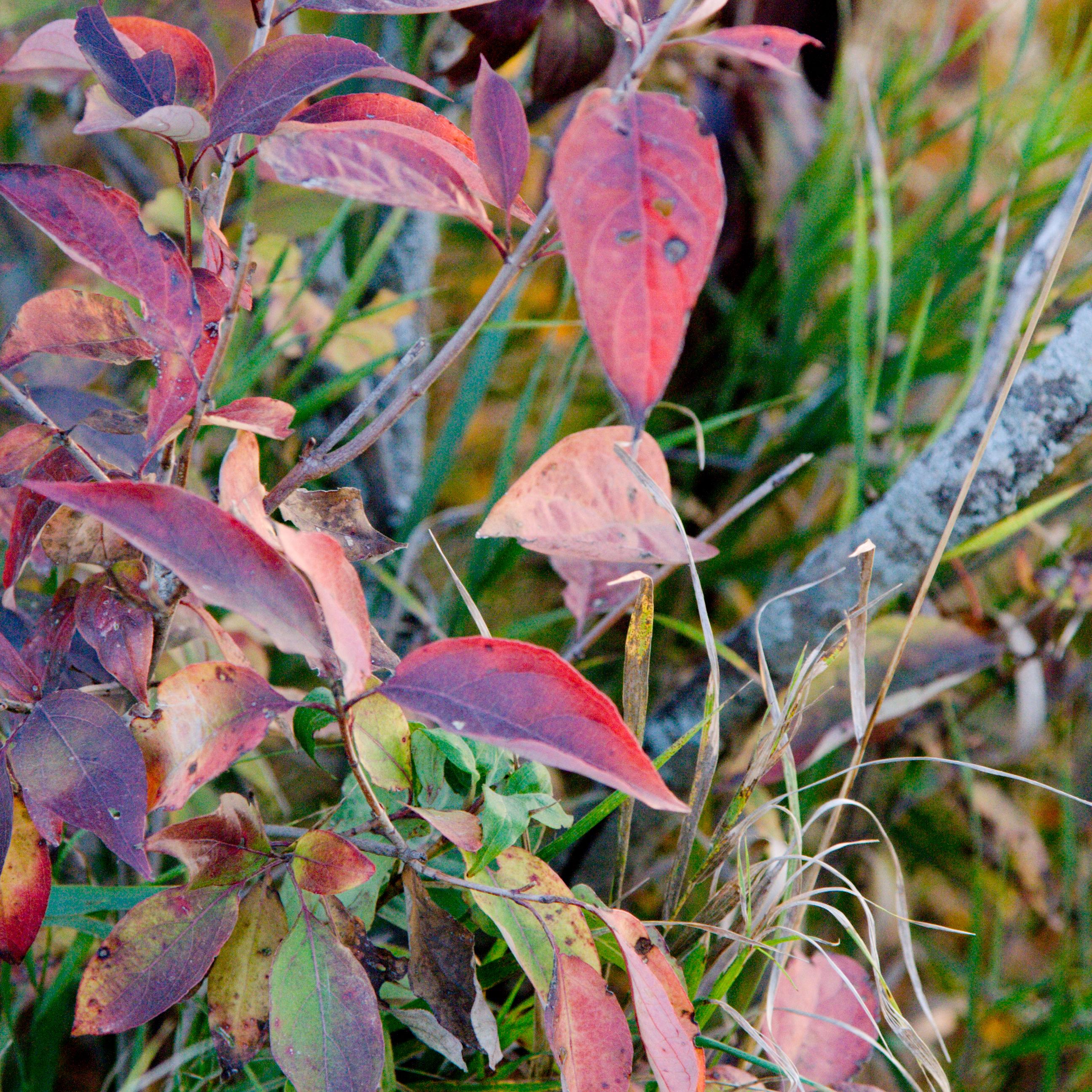}&
\includegraphics[width=0.17\linewidth]{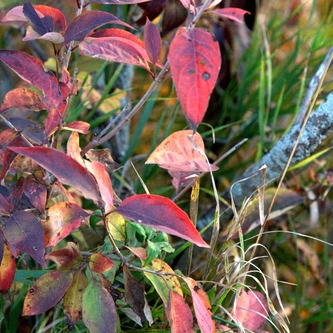}&
\includegraphics[width=0.17\linewidth]{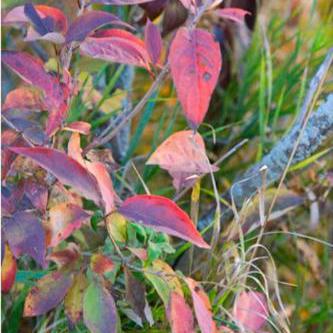}&
\includegraphics[width=0.17\linewidth]{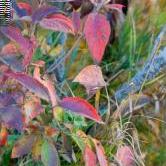}&
\includegraphics[width=0.17\linewidth]{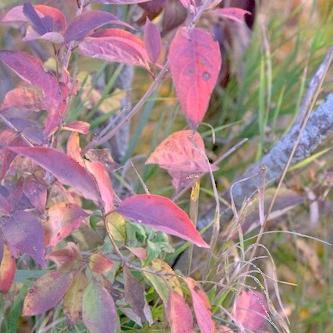}\\
Raw&Human&Ours&Yan~\etal&Pix2Pix&\hspace{-1mm}Hwang~\etal\\
\end{tabular}
\end{adjustbox}
\end{center}
\vspace{-2mm}
\caption{Qualitative comparison with Yan~\etal\cite{Yan2016}, Pix2Pix~\cite{Pix2Pix}, Hwang~\etal\cite{Hwang2012} on \textsc{RANDOM 250}. All approaches are trained with input-retouched pairs by Expert C in MIT-Adobe FiveK. Images for Yan~\etal\cite{Yan2016} and Hwang~\etal\cite{Hwang2012} are taken from the original paper and provided by the original author, respectively.}
\label{fig:comp1}
\vspace{-4mm}
\end{figure*}

\subsection{Input-Retouched Dataset}
\label{sec:exp_raw_retouched}
We compare our method with the previous methods~\cite{Hwang2012,Yan2016} by learning a specific human expert's color adjustment style.
Training setup is the same as Section~\ref{sec:exp_feature}.

Table~\ref{table:comparison1} summarizes the results of different methods and Figure~\ref{fig:comp1} shows example results for qualitative comparison.
In Table~\ref{table:comparison1}, our method performs better than \cite{Hwang2012} and is comparable to \cite{Pix2Pix,Yan2016}. Pix2Pix~\cite{Pix2Pix} is also competitive with \cite{Yan2016}.
Pixel-level prediction methods, such as \cite{Pix2Pix,Yan2016},  performs well in terms of mean $\mathcal{L}^2$ error because local color enhancement provides much flexible mapping while our method is mainly limited by the predefined actions in Table~\ref{table:actions}.
However, pixel-level prediction methods suffer from local artifacts especially in handling high-resolution images. To quantitatively compare these artifacts, we have measured SSIM scores in both Table~\ref{table:comparison1} and Table~\ref{table:distorted_retouched_comp1}, and the scores demonstrate that the superior image quality of our results.
Also, compared to the network directly estimating the pixel values, our method is \textit{interpretable} since our agent actively chooses and applies a sequence of human-defined editing operations. These intermediate retouching actions provide better understanding of the retouching process and can be useful in many scenarios like user-guided photo retouching and personalized tutorial generation.
In addition, our DRL approach outperforms other methods in the proposed \textit{distort-and-recover} scheme, which will be addressed in the following Section~\ref{sec:exp_distort_and_recover}.

\subsection{Distort-and-Recover Training}
\label{sec:exp_distort_and_recover}
We perform an experiment to evaluate our \textit{distort-and-recover} training scheme as described in Section~\ref{sec:data}. In this experiment, we assume that we only have high-quality reference images, not input-retouched pairs. Instead, we distort reference images by applying random actions and use the distorted-reference pairs for training. 


\begin{table}[t]
\begin{center}
\setlength{\tabcolsep}{1.7pt}\small
\begin{tabular}{|l|c|c|}
\hline
Method&mean $\mathcal{L}^2$ error&SSIM\\\hline\hline
Input image&17.07&-\\
Pix2Pix~\cite{Pix2Pix} - no augmentation&14.46&0.699\\
Pix2Pix~\cite{Pix2Pix} - augmented 10 times&13.79&0.804\\
Pix2Pix~\cite{Pix2Pix} - augmented 20 times&13.59&0.786\\
Pix2Pix~\cite{Pix2Pix} - augmented 30 times&13.60&0.772\\
Ours - no augmentation&\textbf{12.15}&\textbf{0.910}\\\hline
\end{tabular}
\end{center}
\caption{Mean $\mathcal{L}^2$ error and SSIM of different approaches trained with the \textit{distort-and-recover} scheme. Retouched images by Expert C on MIT-Adobe FiveK are used as reference images.}
\label{table:distorted_retouched_comp1}
\vspace{-5mm}
\end{table}

\paragraph{Comparison with supervised learning}
We use the images retouched by Expert C in MIT-Adobe FiveK as reference images and distort them as inputs to train our agent. We also train Pix2Pix~\cite{Pix2Pix} with the same \textit{distort-and-recover} training scheme.
Table~\ref{table:distorted_retouched_comp1} shows the comparison result. At first, we train our agent with 4,750 pairs of distorted-reference images where only one randomly distorted image is generated for each reference image. Our approach shows decent performance with this setting, although the color distribution of raw test images are never shown to the agent network. Examples of distorted images are included in the supplementary material. In contrast, the performance of Pix2Pix drops severely, compared to the previous result with input-retouched pairs in Table~\ref{table:comparison1}. When the large number of augmented pairs, in which a reference image is associated with multiple distorted images, are used for training, the performance of Pix2Pix gets better but is quickly saturated.


This experiment shows that our DRL approach works well with the \textit{distort-and-recover} training scheme even without extra data augmentation, while supervised learning methods may suffer from unknown source color distributions. 
\JC{We conjecture the high performance of our approach is due to the efficient state-exploring nature of reinforcement learning:
In the DQN framework, the search space is constrained with the discretized action set, and Markov property enables the agent to efficiently search the near color state.}
On the other hand, Pix2Pix can be seen as one-step action that translates input distribution to reference distribution with a high non-linearity in a large search space. Its performance depends highly on the training set.
That is, if a training set covers limited input distributions, then the trained network suffers from poor generalization in test cases.
Therefore, our DRL agent is robust to the input distribution shift while Pix2Pix is not.

\paragraph{Learning a style filter}
We perform an experiment to learn certain styles with the \textit{distort-and-recover} scheme. We apply a predefined filter (Nashville in Instagram) to the retouched images by Expert C to create a set of reference images in a new style. We train our DRL agent with distorted and Nashville-filtered image pairs and test the agent with the raw input images on \textsc{Random~250}. The mean $\mathcal{L}^2$ error between the inputs and corresponding ground truth images is $27.78$, while the mean error between our outputs and the groundtruth images is $17.87$. \JC{Qualitative results are shown in the supplementary material.} These results verifies that our agent is able to learn difference retouching styles only with desired style images, demonstrating the possibility of personalized enhancement from a user-curated dataset with our \textit{distort-and-recover} training scheme.

\begin{figure}
\begin{center}
\setlength{\tabcolsep}{1pt}\small
\begin{tabular}{ccccccc}
\includegraphics[width=0.13\linewidth]{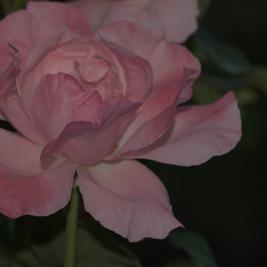}&
\includegraphics[width=0.13\linewidth]{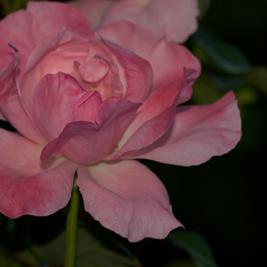}&
\includegraphics[width=0.13\linewidth]{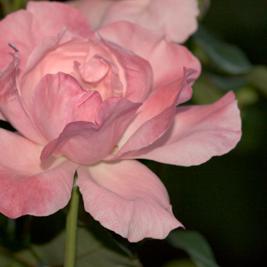}&
\includegraphics[width=0.13\linewidth]{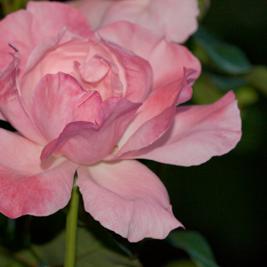}&
\includegraphics[width=0.13\linewidth]{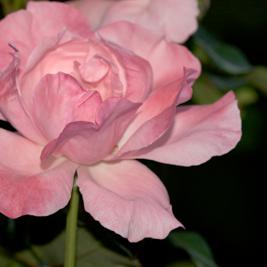}&
\includegraphics[width=0.13\linewidth]{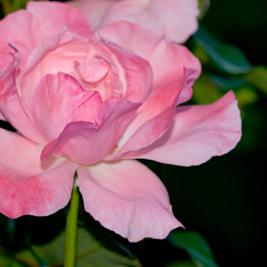}&
\includegraphics[width=0.13\linewidth]{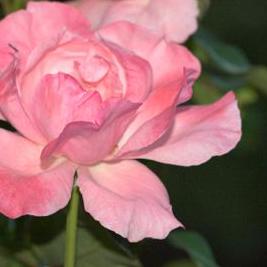}\\
\includegraphics[width=0.13\linewidth]{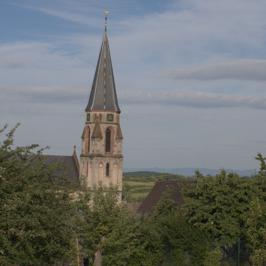}&
\includegraphics[width=0.13\linewidth]{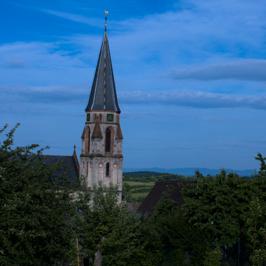}&
\includegraphics[width=0.13\linewidth]{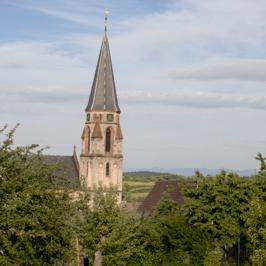}&
\includegraphics[width=0.13\linewidth]{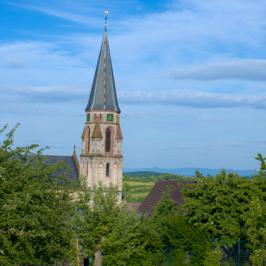}&
\includegraphics[width=0.13\linewidth]{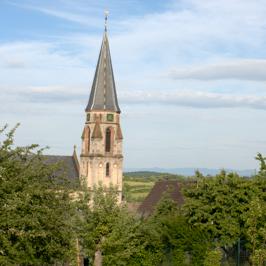}&
\includegraphics[width=0.13\linewidth]{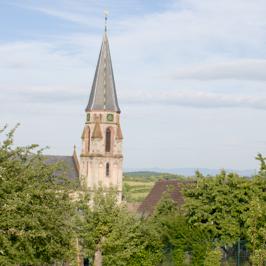}&
\includegraphics[width=0.13\linewidth]{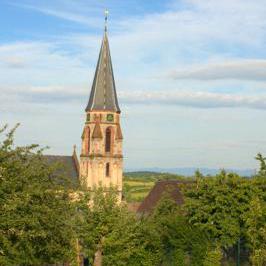}\\
Input&A&B&C&D&E&Ours\\
\end{tabular}
\end{center}
\vspace{-3mm}
\caption{Qualitative comparison between five human experts' retouching (A,B,C,D,E) and our result on MIT-Adobe FiveK. We train our agent with 150K reference images from Shutterstock using the \textit{distort-and-recover} training scheme.}
\label{fig:shutterstock1}
\vspace{-3mm}
\end{figure}

\paragraph{Distort-and-Recover with Shutterstock 150K}
To build an automatic color enhancement system that handles a wide range of image content, it is essential to have a large number of training data with diverse context. Our \textit{distort-and-recover} scheme gives a practical solution to this problem. As an example case study, we collect high-quality reference images from a stock image website, Shutterstock\footnote{https://www.shutterstock.com/}. We have crawled 150K thumbnail images from Shutterstock by searching general keywords of ``food, flower, nature, portrait, mountain", sorted by popularity. Then, we train our DRL agent with the crawled reference images. The crawled images are retouched or curated by various photographers, so the images have no specific style or personal preference. From such reference images, we can expect that the agent network learn a generally pleasing style according to image contents. We test our agent network for the raw images on MIT-Adobe FiveK as inputs. Figure~\ref{fig:shutterstock1} shows examples of our enhancement results compared with human experts.

We evaluate the performance of our method through a user study. For the user study, we compare the retouching results from our agent with four different sets: 1) raw input images as a baseline, 2) Pix2Pix as one of state-of-the-art supervised learning methods, 3) \textit{Lightroom} -- a commercial photo retouching software, 4) expert-retouched images as groundtruth. During training the Pix2Pix model, we observed that the model diverges and produces poor results with a large number of distorted-reference pairs, so we trained Pix2Pix with 10K distorted-reference image pairs. For Lightroom, we have retouched images by applying both `auto white-balance' and `auto-tone adjustment' functions in Adobe Lightroom. We prepared `expert-retouched' images by randomly selecting one expert result out of five corresponding expert-retouched images for each input.

The user study is conducted on the Google Forms platform~\footnote{https://docs.google.com/forms/}. For each question, randomly shuffled five images are shown to the respondents. Respondents are told to give scores ranging from 0 (worst) to 5 (best) for each given image. There are 50 question sets of images, 30 respondents, and a total of 7500 scores are rated. We recruited respondents from a college campus and paid them accordingly. All the images used in the survey are randomly selected from \textsc{RANDOM250} with other corresponding results. The results are listed in the supplementary materials.

Table~\ref{tab:userstudy} shows the result of the user study.
We report `mean scores' and `normalized mean scores' with standard deviations. Since each respondent has different sense of score scale due to the subject manner of a problem, in `normalized mean scores', we normalize scores per respondent so that each respondent's answers have zero mean and unit variance distribution. In the user study, our method significantly outperformed Pix2Pix and also shows superior performance than the results of Lightroom, one of the best commercial retouching software which is highly optimized for photo retouching application. 

Compared with the quantitative evaluation using input-retouched pairs, Pix2Pix produced poor results and got low scores in this user study, while our method works robustly in both cases. We note that we used distorted images for training and raw images for testing in our user study, therefore we conjecture that the performance drop of Pix2Pix comes from the gap between color distributions of training and test images. More specifically, in the input-retouched pairs, the test input distributions and the train input distributions are expected to be similar, so fully supervised networks are able to learn mappings from input distributions to target distributions; in the distorted-reference pairs, however, training inputs are generated with random distortions, therefore networks should handle mappings from unknown distributions to target distributions. We empirically observe that our method performs well under such circumstances and we argue that it is due to the state-exploring nature of deep reinforcement learning. 



\begin{table}
	\begin{center}
		\setlength{\tabcolsep}{1.6pt}\small
		\begin{tabular}{|l|c|c|}\hline
			Retouch method& Mean score & Mean score (normalized)\\\hline\hline
			Input image & 2.11 (\(\pm\)1.20) & -0.43 (\(\pm\)0.90)\\\hline
			Pix2Pix~\cite{Pix2Pix} & 2.09 (\(\pm\)1.31) & -0.44 (\(\pm\)0.97)\\\hline
			Lightroom & 2.74 (\(\pm\)1.31) & 0.09 (\(\pm\)0.93)\\\hline
			\textbf{Ours} & \textbf{2.86} (\(\pm\)1.22) & \textbf{0.19} (\(\pm\)0.88)\\\hline\hline
			Expert & 3.37 (\(\pm\)1.28) & 0.60 (\(\pm\)0.92)\\\hline
		\end{tabular}
	\end{center}
	\caption{\textbf{User study results.} Pix2Pix~\cite{Pix2Pix} and ours is trained with Shutterstock 150K images using the \textit{distort-and-recover} training scheme, and the test images are input images from MIT-Adobe FiveK dataset. The \textit{normalized} scores are computed after normalizing scores to zero mean and unit variance for each respondent.}
	\label{tab:userstudy}
	\vspace{-5mm}
\end{table} 

\section{Conclusion}
\label{sec:conclusion}
We have proposed a novel method for color enhancement. With the deep reinforcement learning approach, we have explicitly modeled a human retouching process and estimated interpretable retouching steps from predefined actions without extra annotations. To resolve the difficulty of collecting training data, we also have presented the \textit{distort-and-recover} training scheme, which enables us to learn our model with large-scale data in an economic way. We have shown that our DRL approach compensates the lack of generality in distorted images effectively and is generalized well to unknown input distribution in test time. Our DRL-based color enhancement with data efficient training scheme has shown a new possibility in interesting applications including automatic tutorial generation, and personalized enhancement, and we hope to explore these applications in the future.

\section*{Acknowledgement}
This work was supported by the Technology Innovation Program (No. 10048320), funded by the Ministry of Trade, Industry \& Energy (MI, Korea).



{\small
\bibliographystyle{ieee}

}

\end{document}